\documentclass[runningheads]{llncs}

\usepackage[T1]{fontenc}
\usepackage[utf8]{inputenc}
\usepackage{graphicx}
\usepackage{booktabs}
\usepackage{multirow}
\usepackage{url}

\usepackage{xspace}
\usepackage{amsmath}
\usepackage{tikz}
\usetikzlibrary{shapes.geometric,positioning,arrows.meta}
\usepackage{float}
\usepackage[hidelinks]{hyperref}

\newcommand{\method}{SchemaShield-Lite\xspace}
\newcommand{\artifactrepo}{https://github.com/AniruddhaChattopadhyay/tool-drift}
\newcommand{\artifactbase}{\artifactrepo/blob/main}
\newcommand{\artifactrepolink}{\href{\artifactrepo}{public \texttt{tool-drift} repository}}

\newcommand{\artifactfile}[1]{\href{\artifactbase/#1}{\nolinkurl{#1}}}

\begin{document}
\raggedbottom

\title{Robust Function Calling for LLM Agents Under Interface Drift}
\titlerunning{Robust Function Calling Under Interface Drift}

\author{Aniruddha Chattopadhyay\inst{1} \and
Kanishka Halder\inst{1} \and
Partha Sarathi Roy\inst{1}}
\authorrunning{A. Chattopadhyay et al.}
\institute{Independent Researcher}

\maketitle

\begin{abstract}
Tool-augmented language-model agents rely on interface text: tool descriptions,
argument schemas, and the candidate tool list shown at inference time. These
interfaces often change after deployment even when the underlying tool semantics
remain fixed, creating a practical reliability problem for function-calling
systems. We study this failure mode as \emph{interface drift} and evaluate whether
a lightweight, training-free repair layer can recover robustness without retraining
the base model. Our method, \method, combines schema validation, candidate-set
retry for missing tool names, and one-shot constrained repair. On Qwen3.5-9B, it
raises DICE-300 accuracy from $0.717$ to $0.813$ and BFCL-200 accuracy from
$0.580$ to $0.805$, recovering $19/22$ and $33/45$ drift-induced failures with no
observed repair harms. Naive retry does not explain the gains. Ablations show that
candidate-set drift, especially distractor tools, accounts for most observed
damage, while description and schema edits alone are less harmful. The results
position validation-and-repair as a practical robustness layer for deployed
LLM-agent tool use.
\end{abstract}

\section{Introduction}
Modern language-model agents increasingly depend on external tools for retrieval,
execution, planning, and actuation. In practice, however, tool use is mediated
entirely through text: tool descriptions, parameter schemas, enum values, and
candidate tool lists. Recent work shows that these interface details strongly affect
tool selection and argument generation
even when the tool semantics are unchanged \cite{guo2026rewrite,faghih2025gaming}.
This creates a deployment problem: tool interfaces drift over time as developers
rename parameters, revise descriptions, add examples, or expand candidate tool sets.

We study this problem as \emph{interface drift}: benign, semantics-preserving
changes to tool interfaces that occur after deployment. Our goal is not to train a
new tool-calling model, but to quantify drift sensitivity and test whether a compact
inference-time repair layer can recover reliability without retraining.

We instantiate drift with three deployment-motivated synthetic perturbations: (i)
stale description examples that mention legacy field names, (ii) schema renaming,
and (iii) distractor tools. Our primary evaluation uses Qwen3.5-9B on a
300-example DICE-BENCH slice \cite{jang2025dicebench} and a 200-example BFCL
slice \cite{bfcl2024} under a deployment-realistic non-oracle repair protocol.
Earlier oracle-constrained sweeps on additional open models are retained only as
upper-bound context and for secondary analyses.

This paper makes three empirical contributions, demonstrated primarily on
Qwen3.5-9B with a supporting replication on Qwen3.5-35B-A3B:

\begin{enumerate}
\item \textbf{Non-oracle inference-time repair recovers drift-induced failures on
the primary model.} On Qwen3.5-9B, \method raises DICE-300 accuracy from $0.717$
to $0.813$ and BFCL-200 accuracy from $0.580$ to $0.805$, while showing no
observed repair harms on the originally-correct slice in either benchmark. A
same-protocol BFCL replication on Qwen3.5-35B-A3B ($0.635 \rightarrow 0.845$)
confirms the effect is not confined to a single model size, though the cleanest
non-oracle evidence remains within the Qwen family.

\item \textbf{Naive retry does not explain the gains.} In the separate baseline
runs reported in Section~\ref{sec:naive-retry}, naive retry reaches $0.727$
versus repaired $0.813$ on DICE-300 and $0.610$ versus $0.795$ on BFCL-200.
A second attempt alone therefore cannot explain the recovery.

\item \textbf{Missing tool names are the main non-oracle bottleneck, and most
primary-model drift misses remain validator-visible.} In the non-oracle setting,
adding a candidate-set retry improves repaired accuracy from $0.780$ to $0.813$ on
DICE and from $0.725$ to $0.805$ on BFCL while reducing unresolved repair targets
from $35$ to $15$ and from $34$ to $9$ respectively. In the final primary runs,
only $2/22$ drift misses on the DICE originally-correct slice and $0/45$ on BFCL
remain validator-invisible.
\end{enumerate}

\section{Related Work}
Recent work has established that tool interfaces are not neutral wrappers around
tool use; they are part of the learning and inference problem itself.
Guo et al.~\cite{guo2026rewrite} show that rewriting tool descriptions can
substantially improve agent reliability, especially when candidate tool sets are
large. Faghih et al.~\cite{faghih2025gaming} show the opposite direction:
small edits to tool descriptions can dramatically shift tool preference, exposing a
fragility in description-mediated tool use. Rabinovich and Anaby-Tavor
\cite{rabinovich2025robustness} study robustness in agentic function calling under
toolkit changes and other perturbations, while our focus is narrower: post-deployment
interface drift with fixed underlying tool semantics.

Several recent papers study robustness and recovery in tool-use systems.
Zhang et al.~\cite{zhang2026fission} show that execution-error recovery remains a
major weakness, particularly for smaller models, and propose RL-based training to
improve it. Our setting is narrower and more deployment-oriented: we target
inference-time robustness under interface drift rather than post-training
optimization.

Our repair stack also connects to structured-output enforcement via constrained
decoding. Grammar-constrained and grammar-aligned decoding study how to guarantee
or preserve valid structured outputs during generation
\cite{geng2023grammar,park2024grammar}. We instead study post-hoc validation and
repair after the base model has already produced a drifted tool call.
Feeding error feedback back into the prompt also relates to judge-guided prompt
refinement, which improves reliability without weight updates but has clear
boundary conditions~\cite{chattopadhyay2026metatuning}; in our setting, the
feedback comes from deterministic schema validation rather than an LLM judge.

Benchmark work has also highlighted that realistic function calling is still far from
solved. ComplexFuncBench introduces multi-step and constrained long-context function
calling \cite{zhong2025complexfuncbench}, while DICE-BENCH focuses on multi-round,
multi-party dialogues with tool-related information dispersed across conversation
turns \cite{jang2025dicebench}. Other benchmarks, including API-Bank,
$\tau$-bench, and ToolSandbox, similarly stress planning, state, and interaction
in tool-use settings \cite{li2023apibank,yao2024taubench,lu2024toolsandbox}. We
evaluate on both DICE-BENCH and the Berkeley Function Calling Leaderboard (BFCL)
\cite{bfcl2024} because they capture complementary settings: DICE-BENCH exercises
conversational multi-turn dialogues, while BFCL tests single-turn tool calls across
diverse APIs.

Broader tool-use work has focused on teaching models to call APIs, retrieving
relevant tools, and combining reasoning with action
\cite{schick2023toolformer,patil2023gorilla,qin2023toolllm,yao2022react}. Our
focus is different: robustness when an already-deployed tool interface changes.

This setting also connects to software-engineering work on API evolution and
contract checking. AutoGuard, for example, compares OpenAPI descriptions across
versions and flags changes to methods, parameters, responses, and schemas
\cite{lercher2025autoguard}. Serbout and Pautasso
\cite{serbout2023versioning} study how Web API versioning practices align with the
actual changes made across releases. We borrow the intuition that interface evolution is a
first-class engineering concern, but study its effect on LLM tool use rather than
on conventional API clients.

Compared with dataset-generation work such as APIGen \cite{liu2024apigen}, our
objective is different. We do not create new large-scale tool-calling training data.
Instead, we evaluate whether a lightweight repair layer can stabilize existing models
against a specific class of post-deployment failures.

\section{Problem Formulation}
Let a tool be represented by an interface
$I = (d, s, C)$, where $d$ is the natural-language description, $s$ is the argument
schema, and $C$ is the candidate tool set shown to the model. The underlying tool
semantics are assumed fixed. An \emph{interface drift} operator $\Delta$ maps $I$ to
a new interface $\Delta(I)$ that preserves tool semantics but perturbs surface form.
For clarity, we view $\Delta$ as a composition of three operators,
\[
\Delta = \delta_c \circ \delta_s \circ \delta_d,
\]
where $\delta_d$ edits only the description text $d$, $\delta_s$ edits only the
schema surface form $s$, and $\delta_c$ edits only the visible candidate set $C$.
Each operator is semantics-preserving with respect to the benchmark gold tool call;
the evaluation question is whether the model remains robust to these surface-form
changes at inference time.

We study three drift types:
\begin{itemize}
\item \textbf{Description drift}: add stale or legacy examples to the description that
refer to older field names.
\item \textbf{Schema drift}: rename parameters while preserving semantics.
\item \textbf{Candidate-set drift}: add distractor tools with overlapping descriptions.
\end{itemize}

Given user context $x$, tool interface $I$, and model $M$, the base system outputs a
tool call $y = M(x, I)$. We evaluate three stages:
\begin{enumerate}
\item \textbf{Original}: inference on the original tool interface.
\item \textbf{Drifted}: inference on the drifted interface $\Delta(I)$.
\item \textbf{Repaired}: inference-time repair of an invalid or incomplete drifted call.
\end{enumerate}

Our central question is whether an inference-time repair layer can recover the
performance lost from Original to Drifted without harming examples that were already
correct.

\section{Method}
\label{sec:method}
\subsection{\method}
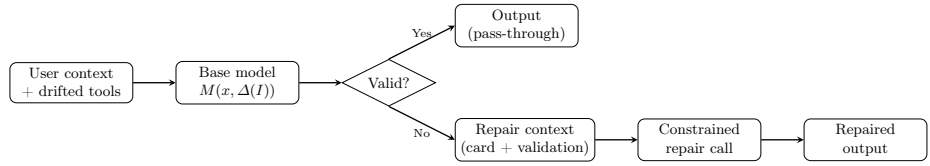
\begin{figure}[t]
\centering
\resizebox{\linewidth}{!}{%
\begin{tikzpicture}[
    node distance=0.9cm and 0.9cm,
    box/.style={draw, rounded corners, minimum width=2.6cm, minimum height=0.8cm, align=center, font=\small},
    decision/.style={draw, diamond, aspect=2, minimum width=1.4cm, align=center, font=\small},
    arrow/.style={->, >=stealth, thick},
]
\node[box] (input) {User context\\+ drifted tools};
\node[box, right=of input] (model) {Base model\\$M(x, \Delta(I))$};
\node[decision, right=of model] (valid) {Valid?};
\node[box, above right=0.5cm and 0.9cm of valid] (pass) {Output\\(pass-through)};
\node[box, below right=0.5cm and 0.9cm of valid] (repair) {Repair context\\(card + validation)};
\node[box, right=of repair] (repairmodel) {Constrained\\repair call};
\node[box, right=of repairmodel] (output) {Repaired\\output};

\draw[arrow] (input) -- (model);
\draw[arrow] (model) -- (valid);
\draw[arrow] (valid.north) -- node[above, font=\scriptsize] {Yes} (pass.west);
\draw[arrow] (valid.south) -- node[below, font=\scriptsize] {No} (repair.west);
\draw[arrow] (repair) -- (repairmodel);
\draw[arrow] (repairmodel) -- (output);
\end{tikzpicture}
}
\caption{\method pipeline. Invalid tool calls trigger structured validation and
constrained repair; missing tool names are first resolved via candidate-set retry.}
\label{fig:pipeline}
\end{figure}

\method is a lightweight inference-time robustness layer with three parts.
Figure~\ref{fig:pipeline} sketches the full pipeline.

\paragraph{Canonical tool card.}
Each tool is rendered into a compact canonical form containing the tool name,
description, required fields, and field-level type information. This normalizes
interface presentation across tools and emphasizes current field names.

\paragraph{Validation.}
After the base model emits a tool call, a validator checks tool name, required
fields, and argument structure. Validation produces a structured error report rather
than a binary failure signal.

\paragraph{One-shot repair.}
If the drifted call is invalid, the model receives a repair prompt containing the
task context, canonical tool card, invalid call, and validation errors. In the
primary non-oracle protocol, repair proceeds in two stages. If the base call names
a visible tool, validation runs against that tool and a constrained repair call is
issued for that same visible tool. If the base call omits the tool name, we first
run a candidate-set retry over the visible tools to produce a concrete tool call;
once a visible tool is resolved, a second repair call may be constrained to that
selected tool. Where provider support allows it, this second step uses
\emph{forced-tool repair}; otherwise we fall back to unconstrained tool calling or
JSON repair.

As our component ablation in Section~\ref{sec:component-ablation} shows, it is the
combination of structured validation errors and tool-constrained repair that does
the heavy lifting once a visible target tool has been resolved. The canonical card
is useful at the initial prediction step as drift-prevention prompting, but
contributes negligibly to repair-stage recovery in our oracle-constrained
upper-bound ablations.

\subsection{Repair Prompting Details}
Three implementation details turned out to matter materially:
\begin{itemize}
\item The repair prompt explicitly instructs the model to infer all required fields
from the dialogue even when the invalid call is empty.
\item When the invalid call omits the tool name, a candidate-set retry over the
visible tools is needed before tool-constrained repair can fire.
\item The repair stage uses a larger token budget than the base prediction stage.
Raising repair capacity from the earlier default to $512$ tokens fixed truncated
repair outputs on longer DICE dialogues.
\end{itemize}

\subsection{Per-Example Evaluation Loop}
\label{subsec:eval-loop}

Each benchmark example is evaluated in three stages: Original, Drifted, and
Repaired. Drift is injected once, the gold call is re-expressed in the renamed
schema before scoring, and repair fires only when validation flags the drifted
call as invalid. We average the three per-example correctness indicators over
the benchmark slice to obtain the headline accuracies.

\section{Experimental Setup}
\subsection{Benchmarks and Data Slices}
We evaluate on two benchmarks. For DICE-BENCH \cite{jang2025dicebench}, we use a
300-example subset drawn from single-function round-1 dialogues (allowing duplicate
functions to reach sufficient scale). For BFCL \cite{bfcl2024}, we use a 200-example
subset drawn in equal parts from four categories: simple Python, multiple-function,
live-simple, and live-multiple.

\subsection{Models}
We evaluate four open instruction-tuned models accessed via OpenRouter:

\begin{itemize}
\item \textbf{Qwen3.5-9B}: our primary model.
\item \textbf{Qwen3.5-35B-A3B}: larger MoE variant.
\item \textbf{Llama-4-Scout}: Meta's 2025 open model.
\item \textbf{Mistral-Small-3.2-24B}: recent Mistral dense model.
\end{itemize}

All experiments use temperature $0.0$ for low-variance evaluation, though hosted
inference may still exhibit provider-side nondeterminism.

\subsection{Drift Configurations}
For DICE main runs, we use \emph{mild} severity: two description modes
(legacy-example and marketing), two schema modes
(rename-parameters and reorder-parameters), and one distractor
tool. For BFCL we use \emph{medium} severity: three description modes, three schema
modes, and three distractors, reflecting a more aggressive drift scenario. These
operators are author-designed and deployment-motivated rather than mined from real
API changelogs; we return to that limitation in Section~\ref{sec:limitations}.
To sanity-check ecological validity, we manually audited $11$ official changelog
items from Notion and Stripe spanning 2022--2026. Six map directly to our
documentation/example and parameter/schema operators, three more are close
candidate-set analogues where vendors add or consolidate neighboring tools or
endpoints, and two are larger structural refactors outside our
semantics-preserving scope
\cite{notionchangelog2026,stripecheckout2022,stripedefaultvalue2025,stripeinvoicepreview2025,stripedelaydays2025}. We
therefore treat the synthetic operators as a compact model of common surface-form
churn rather than as a measured distribution over all API evolution events.

\subsection{Repair Protocols}
We use two repair protocols. Our \emph{primary} protocol is non-oracle and is the
main evidence in this paper: the validator checks the model-selected visible tool,
and when the model omits the tool name, a candidate-set retry over only the visible
tools resolves a target before repair. For broader model coverage and component
ablations, we also retain earlier \emph{oracle-constrained upper-bound} sweeps in
which the repair stage is given the benchmark target tool. These upper-bound runs
remain useful for cross-model context and ablation, but they are not the paper's
deployment-facing evidence.

\subsection{Metrics}
We report:
\begin{itemize}
\item \textbf{Original score}: accuracy on the unmodified interface.
\item \textbf{Drifted score}: accuracy under interface drift.
\item \textbf{Repaired score}: accuracy after one-shot repair.
\item \textbf{Naive retry score}: accuracy under a second identical drifted call
(no canonical card, no validation feedback). This baseline isolates the value of
simply retrying.
\item \textbf{Originally-correct slice}: the subset of examples that the model solves
correctly before drift. This slice isolates \emph{true drift harm} from baseline
errors.
\item \textbf{Repair recoveries}: drift-induced failures recovered by repair.
\item \textbf{Repair harms}: drifted-correct examples broken by repair.
\item \textbf{Repair trigger rate}: fraction of examples for which the validator
invokes the repair step.
\end{itemize}

\subsection{Reproducibility and Release}
All reported results come from fixed benchmark slices, fixed drift configurations,
and fixed random seeds. The per-run JSON artifacts store example IDs, prompts,
drifted tool schemas, validation outputs, repair targets, repair strategy, and
token usage, and our evaluation uses the normalized semantic matching policy
reported in each run summary. We release the experiment code, drift-generation
scripts, prompts, configs, subset definitions, scoring/validation logic, and
raw run artifacts in a public repository (\artifactrepolink). The artifact package
stores example-level traces for the main success and limitation modes, including
prompts, validation errors, repaired calls, and gold calls.

\section{Results}
\subsection{Primary Non-Oracle Results}
Table~\ref{tab:main-results} summarizes our primary non-oracle Qwen3.5-9B results
on DICE-300 and BFCL-200, together with oracle-constrained upper bounds for
reference only. Under the non-oracle candidate-retry protocol, \method improves
over the drifted baseline on both benchmarks with no observed repair harms on the
originally-correct slice.

\begin{table}[t]
\caption{Primary non-oracle Qwen3.5-9B results on DICE-300 and BFCL-200, with
oracle upper bounds. Brackets give 95\% bootstrap CIs (seed~42).}
\label{tab:main-results}
\centering
\scriptsize
\setlength{\tabcolsep}{3.5pt}
\resizebox{\linewidth}{!}{%
\begin{tabular}{llccccccc}
\toprule
Benchmark & Protocol & Orig. [95\% CI] & Drift. [95\% CI] & Repair [95\% CI] & OC & Rec/Miss & Harms & Trig\% \\
\midrule
\multirow{2}{*}{DICE-300}
& Non-oracle + candidate retry & 0.757 [0.70, 0.80] & 0.717 [0.67, 0.77] & \textbf{0.813 [0.77, 0.86]} & 227 & 19/22 & 0 & 16.0 \\
& Oracle upper bound           & 0.750 [0.70, 0.80] & 0.713 [0.66, 0.76] & \textbf{0.827 [0.79, 0.87]} & 225 & 17/18 & 0 & 21.0 \\
\midrule
\multirow{2}{*}{BFCL-200}
& Non-oracle + candidate retry & 0.790 [0.74, 0.84] & 0.580 [0.51, 0.65] & \textbf{0.805 [0.75, 0.86]} & 158 & 33/45 & 0 & 32.0 \\
& Oracle upper bound           & 0.790 [0.74, 0.84] & 0.640 [0.58, 0.71] & \textbf{0.795 [0.74, 0.85]} & 158 & 23/35 & 0 & 31.0 \\
\bottomrule
\end{tabular}
}
\par\smallskip\raggedright\scriptsize
OC\,=\,originally-correct count; Rec\,=\,recoveries on drift-induced failures;
Harms\,=\,repair harms on originally-correct slice; Trig\%\,=\,repair trigger rate.
Rows come from separate runs, so Original scores may differ slightly across protocols.
\end{table}

The primary non-oracle results are strong on both benchmarks. On DICE-300, repair
improves Qwen3.5-9B from $0.717$ to $0.813$ and recovers $19/22$ drift-induced
failures on the originally-correct slice. On BFCL-200, repair improves from $0.580$
to $0.805$ and recovers $33/45$ drift-induced failures. The remaining gap to the
oracle upper bound is small on DICE ($0.813$ vs $0.827$), and on BFCL the final
non-oracle candidate-retry run slightly exceeds the earlier oracle-style run
($0.805$ vs $0.795$).

We therefore focus our interpretation on the large repaired-vs-drifted gaps, not on
small differences between nearby runs. The bootstrap intervals in
Table~\ref{tab:main-results} support the main claim that structured repair helps,
but they do not justify strong claims about minor oracle-vs-non-oracle differences.
Paired exact McNemar tests on the final non-oracle runs also support the main
repair effect: repaired beats drifted with $p=3.7\times 10^{-9}$ on DICE and
$p=5.7\times 10^{-14}$ on BFCL. On shared examples, the repaired-score difference
between the oracle upper bound and the final non-oracle protocol is itself not
significant on either benchmark (DICE exact McNemar $p=0.125$; BFCL
$p=0.754$), which is consistent with the small raw score gaps in
Table~\ref{tab:main-results}.

The candidate-set retry is load-bearing in the non-oracle setting. Relative to an
earlier non-oracle run that simply skipped unresolved tool names, candidate retry
raises repaired accuracy from $0.780$ to $0.813$ on DICE and from $0.725$ to
$0.805$ on BFCL, while reducing unresolved repair targets from $35$ to $15$ and
from $34$ to $9$ respectively.

\paragraph{Oracle upper bounds for context.}
The oracle rows in Table~\ref{tab:main-results} remain useful as upper-bound
context, but they are not the paper's main evidence. We use them to localize
mechanisms and portability limits, not to define the deployment-facing claim.

\paragraph{Repair cost overhead.}
On the final non-oracle Qwen3.5-9B runs, \method triggers repair on $48/300$
examples ($16.0\%$) in DICE and $64/200$ examples ($32.0\%$) in BFCL. Total repair
tokens are $103{,}221$ against $316{,}185$ drifted tokens on DICE ($32.6\%$
relative overhead) and $154{,}249$ against $327{,}289$ drifted tokens on BFCL
($47.1\%$ relative overhead). Because repair is gated by validation, this overhead
is paid only when the drifted call is invalid or incomplete. On a refreshed
BFCL-200 live rerun of the same non-oracle protocol with latency instrumentation,
average original-call latency is $8.8$s, average drifted-call latency is $9.3$s,
and average repair-call latency when repair fires is $13.0$s. With a $27.0\%$
repair trigger rate on that rerun, this corresponds to roughly $3.5$s expected
added latency per example.

\subsection{Baselines Do Not Explain the Gains}
\label{sec:naive-retry}
A first-order alternative explanation is that repair works simply because the model
gets a second attempt. We test this with two baselines
(Table~\ref{tab:naive-retry}). \emph{Naive retry} re-runs the identical drifted
prompt with no new information. \emph{Tool-list re-prompt} tells the model the
call was invalid and lists the available tool names, but provides no canonical
card, no validation errors, and no view of the invalid call. This second baseline
isolates whether simply flagging the error and listing tools explains the gains.

Neither baseline approaches \method. Naive retry reaches $0.727$ on DICE and
$0.610$ on BFCL; tool-list re-prompt improves only marginally to $0.753$ and
$0.625$ respectively. Both remain far below \method ($0.813$ and $0.795$).
The structured validation feedback, not merely knowing which tools are
available, is therefore the load-bearing component. The paired
repaired-vs-naive contrast is highly significant on both benchmarks
(exact McNemar $p=4.1\times 10^{-9}$ on DICE; $p=1.1\times 10^{-13}$ on BFCL).

\begin{table}[t]
\caption{Baselines vs \method on Qwen3.5-9B (non-oracle). Scores from separate
runs; minor differences from Table~\ref{tab:main-results} reflect run-level variance.}
\label{tab:naive-retry}
\centering
\small
\begin{tabular}{lccccc}
\toprule
Benchmark & Drifted & Naive Retry & Tool-List Re-prompt & \method & $\Delta$ over Drifted \\
\midrule
DICE-300   & 0.713 & 0.727 & 0.753 & \textbf{0.813} & $+0.100$ \\
BFCL-200   & 0.590 & 0.610 & 0.625 & \textbf{0.795} & $+0.205$ \\
\bottomrule
\end{tabular}
\end{table}

\subsection{Repair Coverage: Drift Recovery Dominates and Most Misses Are Validator-Visible}
\label{sec:repair-coverage}
The ``repair $>$ original'' effect is real, but it does not dominate the main
story. Table~\ref{tab:gain-decomposition} decomposes every repaired improvement over
the drifted call into either \emph{drift recovery} on the originally-correct slice
or \emph{baseline patching} on examples the base model already missed before drift.
Most repaired gains come from genuine drift recovery: $19/29$ on DICE and
$33/45$ on BFCL. Baseline patching is material, especially on BFCL, and should be
read as a secondary benefit rather than the paper's central causal claim.

\begin{table}[t]
\caption{Decomposition of repaired improvements on the primary non-oracle
Qwen3.5-9B runs.}
\label{tab:gain-decomposition}
\centering
\small
\begin{tabular}{lcccc}
\toprule
Benchmark & Total gains & Drift recovery & Baseline patching & Harms \\
\midrule
DICE-300 & 29 & 19 & 10 & 0 \\
BFCL-200 & 45 & 33 & 12 & 0 \\
\bottomrule
\end{tabular}
\end{table}

These runs also reveal whether drift misses actually enter the repair path.
On the originally-correct slice, $20/22$ DICE drift misses and all $45/45$ BFCL
drift misses are validator-visible invalid or incomplete calls. Only two DICE
misses remain validator-invisible valid-but-wrong outputs, and BFCL has none.
Across the full benchmark slices, $63/85$ DICE drift misses and $73/84$ BFCL
drift misses are validator-visible. The repair stack therefore addresses most
observed drift failures; valid-but-wrong calls that bypass validation are a
residual rather than dominant failure mode.

\subsection{Drift-Type Ablation: Candidate-Set Drift Dominates}
\label{sec:drift-ablation}
Table~\ref{tab:drift-ablation} decomposes the drift pipeline into its three
components and evaluates each in isolation at DICE-300 on Qwen3.5-9B. Description
drift alone causes only a $0.013$ point accuracy drop. Schema drift alone causes
only $0.007$. By contrast, candidate-set drift (adding a distractor tool) causes a
$0.047$ drop, matching the full combined drift scenario ($0.050$). Because this
ablation is about where the damage comes from, the main interpretation still rests
on the Original-to-Drifted delta. For the repair column, however, only the
distractor-bearing rows require fresh non-oracle evidence. In our DICE-300 slice,
every baseline task has exactly one candidate tool, so under predicted-tool repair
the description-only and schema-only settings collapse to the same effective repair
target as the oracle-target protocol via single-candidate fallback. The non-oracle
question therefore matters only once distractor tools are introduced. A fresh
candidate-set-only non-oracle rerun reaches $0.817$, while the primary DICE main
run already supplies the matching all-combined non-oracle row at $0.813$. The
qualitative conclusion is unchanged: candidate-set drift remains the dominant
source of damage, and non-oracle repair stays close to the earlier oracle upper
bounds.

\begin{table}[t]
\caption{Drift-type ablation on DICE-300 with Qwen3.5-9B. Single-candidate tasks
make description/schema-only repair equivalent to oracle-target.}
\label{tab:drift-ablation}
\centering
\small
\begin{tabular}{lcccr}
\toprule
Drift type & Orig. & Drift. & Repair & Drift Drop \\
\midrule
Description only     & 0.760 & 0.747 & 0.813 & $-0.013$ \\
Schema only          & 0.767 & 0.760 & 0.830 & $-0.007$ \\
Candidate-set only   & 0.757 & 0.710 & 0.817 & $-0.047$ \\
\textbf{All combined}& 0.763 & 0.713 & 0.813 & $-0.050$ \\
\bottomrule
\end{tabular}
\end{table}

This result qualifies a common framing in the interface-drift literature. On this
Qwen3.5-9B DICE-300 ablation, minor description rewording and
semantics-preserving parameter renames do not meaningfully degrade accuracy; the
model is evidently robust to this class of surface perturbation. The observed
failure mode is \emph{candidate-set confusion}: when a distractor tool with an
overlapping description is added to the candidate set, tool selection degrades.
In this slice, the practical implication is that tool-catalog growth deserves
closer monitoring than minor description or schema revisions.

\subsection{Upper-Bound Component Ablation: Validation, Not Canonical Card, Drives Recovery}
\label{sec:component-ablation}
To isolate which parts of the repair stack matter once a target tool has already
been resolved, Table~\ref{tab:component-ablation} evaluates three variants of
\method in an oracle-constrained upper-bound setting on both DICE-300 and BFCL-200
with Qwen3.5-9B: (i) canonical card only, where the card is prepended to the
initial prompt but no validation/repair is applied; (ii) validation-plus-retry,
which uses validation errors in a repair prompt but omits the canonical tool card
from the repair context; and (iii) full \method (card + validation + repair).

\begin{table}[t]
\caption{Oracle-constrained upper-bound component ablation on DICE-300 and
BFCL-200 with Qwen3.5-9B. BFCL uses medium-severity drift, DICE uses mild.}
\label{tab:component-ablation}
\centering
\small
\begin{tabular}{llcccccc}
\toprule
Benchmark & Variant & Drift. & Repair & Rec/Miss & Harms & Trig\% \\
\midrule
\multirow{3}{*}{DICE-300}
& Canonical card only            & 0.770 & 0.770 & 0/9   & 0 & 0.0 \\
& Validation + retry (no card)   & 0.730 & 0.830 & 16/17 & 0 & 20.0 \\
& \textbf{Full \method}          & 0.713 & \textbf{0.827} & 17/18 & 0 & 21.0 \\
\midrule
\multirow{3}{*}{BFCL-200}
& Canonical card only            & 0.810 & 0.810 & 0/11  & 0 & 0.0 \\
& Validation + retry (no card)   & 0.645 & \textbf{0.820} & 26/34 & 0 & 29.5 \\
& Full \method                   & 0.640 & 0.795 & 23/35 & 0 & 31.0 \\
\bottomrule
\end{tabular}
\end{table}

Within this oracle-constrained setting, the BFCL rows confirm and \emph{strengthen}
the DICE finding in three ways.
First, the canonical card at initial-prediction time provides even stronger
drift prevention on BFCL than on DICE: the base drifted score of $0.640$ jumps
to $0.810$, a $17.0$-point drift-prevention effect. Second, the card-only
configuration still never triggers the validator and recovers zero
drift-induced failures, because its valid-but-imperfect outputs pass schema
checking without entering the repair path. Third, validation-plus-retry
without the canonical card at repair time reaches a \emph{higher} repaired
accuracy on BFCL than the full pipeline ($0.820$ vs $0.795$, and $26/34$
recoveries vs $23/35$), matching the pattern observed on DICE where
validation-only nearly ties the full pipeline.

Taken together across both benchmarks, these upper-bound numbers localize the
load-bearing component of \method to the validation error feedback loop
combined with tool-constrained repair, not the canonical tool card presented at
repair time. The card earns its keep as \emph{drift-prevention prompting}
at the initial prediction step, where BFCL shows it can dominate the
remaining drift damage, but can be dropped from the repair prompt without
hurting recovery once the target tool has already been resolved. Because these
ablations are oracle-constrained, we treat them as upper-bound mechanism evidence
rather than as a full explanation of non-oracle portability.

\subsection{Error Analysis}
\label{sec:error-analysis}
On the primary non-oracle Qwen3.5-9B runs, only $3$ DICE examples and $12$ BFCL
examples remain unrecovered on the originally-correct slice. The DICE residuals are
split between one repaired-but-still-mismatched hashtag value and two
validator-invisible value mismatches that never trigger repair. On BFCL, most
unrecovered cases remain validator-visible missing-field failures after repair,
while two cases remain unresolved because the model never supplies a usable tool
name. These patterns are consistent with the main bottleneck story in
Section~\ref{sec:repair-coverage}: the dominant remaining failure modes are either
missing-tool-name resolution or persistent argument completion failures.

Across the oracle-constrained DICE upper-bound sweeps, no repair harms are observed.
However, not every drift miss is recovered. On Qwen3.5-35B, repair recovers $20$ of
$25$ drift misses; on Llama-4-Scout, $12$ of $16$; on Mistral, $8$ of $12$. We
manually inspected every unrecovered case on these three runs and classified them
against the gold call.

The pattern is consistent across models: nearly all unrecovered examples
involve the repair call selecting the correct tool with well-formed arguments
whose string values disagree with gold only by formatting or paraphrase. On
Qwen3.5-35B ($5$ unrecovered), these include one strict whitespace/punctuation
difference (``\texttt{the quick brown fox\ldots}'' vs
``\texttt{The quick brown fox\ldots.}'') and three enum-format or
paraphrase-level mismatches (e.g.\ \texttt{regular\_season} vs
\texttt{regular season}, \texttt{the office} vs \texttt{Office}); a single
genuine failure has an empty arguments object after repair. On Llama-4-Scout
($4$ unrecovered), all four are format-level: \texttt{ISO}-prefixed dates
vs two-field dates (e.g.\ \texttt{2023-11-10} vs \texttt{11-10}), or JSON
with quoted-number string values rather than numeric literals. On Mistral-24B
($4$ unrecovered), the misses include near-synonym topics (\texttt{inspirational}
vs \texttt{inspiration}, \texttt{AI} vs \texttt{artificial intelligence},
\texttt{Google} vs \texttt{GOOGL}) and one wrong numeric amount. Across
13~unrecovered cases in total, only $2$ are genuine inference failures
(empty arguments and one wrong value); the remaining $11$ are strict-match
artifacts rather than true drift failures. The practical implication is that
the paper's reported recovery rates are a lower bound: a lenient evaluator
would count most of the unrecovered cases as correct.

\begin{table}[t]
\caption{Unrecovered oracle-constrained DICE cases on non-primary models.
``Artifact'' = formatting/paraphrase mismatches, semantically correct on inspection.}
\label{tab:unrecovered-summary}
\centering
\small
\begin{tabular}{lccc}
\toprule
Model & Unrecovered & Artifact-like & Genuine failure \\
\midrule
Qwen3.5-35B-A3B & 5 & 4 & 1 \\
Llama-4-Scout   & 4 & 4 & 0 \\
Mistral-24B     & 4 & 3 & 1 \\
\midrule
Total           & 13 & 11 & 2 \\
\bottomrule
\end{tabular}
\end{table}

Mistral-24B exhibits an unusual pattern: its drifted score ($0.777$) slightly
exceeds its original score ($0.740$). Inspection shows that $15/23$ apparent
drift ``gains'' are strict-match artifacts on the original call (formatting
differences in string values) rather than information recovered from the drifted
description. We read this as scorer noise, not evidence that drift helps.

\subsection{Additional Non-Oracle Replications}
\label{sec:additional-replications}
To reduce the single-model dependence of the primary story, we also ran the same
BFCL-200 non-oracle candidate-retry protocol on Qwen3.5-35B-A3B. Repair again
helps materially, lifting accuracy from $0.635$ to $0.845$ (exact McNemar
$p=4.5\times 10^{-13}$). On the originally-correct slice it recovers $33/38$
drift misses, shows no observed repair harms, and leaves only one unresolved
repair target. This does not eliminate same-family dependence, but it shows that
the main BFCL non-oracle effect is not confined to the 9B run.

We also ran the same BFCL-200 non-oracle candidate-retry protocol on
Llama-4-Scout. Repair still improves over the drifted call ($0.635 \rightarrow
0.665$, exact McNemar $p=0.031$) and again shows no observed repair harms on the
originally-correct slice. However, this run exposes an important portability
failure mode: the model omits the tool name frequently, leaving $43$ unresolved
repair targets and limiting recovery on the originally-correct slice to just
$1/11$ drift misses. The Llama run therefore supports some portability of the
repair effect, but also reinforces that non-oracle repair quality depends heavily
on the base model's ability to emit a usable tool name.

We also ran a BFCL-200 non-oracle sanity check on GPT-4o-mini after adding
provider-safe tool-name aliasing for OpenAI-style tool schemas. On this stronger
closed model, drift damage is much smaller to begin with: accuracy falls only from
$0.865$ to $0.840$. Repair lifts the score to $0.860$ with no observed harms and
zero unresolved repair targets, but the gain is small and not significant on this
200-example slice (exact McNemar $p=0.125$). We therefore read the GPT-4o-mini
result as a credibility check rather than as a new headline number: the problem
appears less severe on a stronger closed model, yet the same non-oracle repair
stack does not appear to hurt it.

\section{Discussion}
Our results support three narrower conclusions than the original \method framing
suggested. First, non-oracle inference-time repair is a viable robustness layer for
tool-calling systems under interface drift: on Qwen3.5-9B it improves over drifted
performance on both DICE and BFCL while showing no observed repair harms on the
originally-correct slice. Second, the main non-oracle bottleneck is missing tool
names; a candidate-set retry over the visible tools closes much of the gap to the
oracle upper bound. Third, on our DICE-300 primary-model ablation, candidate-set
drift is a more important failure mode than description or schema drift.

Together these findings simplify the deployment picture. A practitioner does not
need a heavy per-tool canonicalization pipeline; a visible-tool validator, a
candidate-set retry for missing tool names, and a constrained repair call once a
tool is resolved already recover most of the damage we observe. And the monitoring
story is clearer: in our slices, tool-catalog expansion appears more consequential
than minor description or parameter rename churn. The additional BFCL-200
Qwen3.5-35B-A3B replication suggests this practical recipe is not limited to one
exact model size, even if broader cross-family evidence remains thin.

These same results also matter for adversarial interface manipulation. Faghih
et al.~\cite{faghih2025gaming} show that small description edits can steer tool
selection even when semantics are unchanged. Our ablation suggests that the more
serious surface to monitor is not just description wording in isolation, but the
combination of description overlap and candidate-set expansion: once distractor
tools enter the visible set, routing degrades quickly. In that sense, the
non-oracle validator-plus-repair stack is relevant not only to benign interface
drift but also to description-surface attacks that aim to induce tool confusion.

\section{Limitations}
\label{sec:limitations}
Several limitations remain. First, although we now include a strong BFCL
non-oracle replication on Qwen3.5-35B-A3B and a weaker portability check on
Llama-4-Scout, the cleanest successful non-oracle evidence is still concentrated
in the Qwen3.5 family; the broader multi-model results retained in the paper
remain mostly oracle-constrained upper bounds rather than full non-oracle
replications. Second, the BFCL evaluation now covers three open-model non-oracle
runs, but two are Qwen3.5-family results and the third is materially bottlenecked
by missing tool names; we also include one closed-model BFCL sanity check, but
not yet a broader closed-model sweep. Third, our drift operators focus on stale
descriptions, renamed fields, and distractors; other forms of interface drift
(for example, type narrowing, enum replacement, or multi-tool refactors) remain
untested, and although we now add a small manual audit of official vendor
changelogs, the perturbations are still author-designed rather than mined from a
larger empirical change distribution. Fourth, the drift-type and component ablations remain
oracle-constrained upper bounds, so their mechanism-level conclusions may not fully
transfer when non-oracle tool resolution fails. Fifth, evaluation uses hosted
inference and so may incorporate provider-specific tool-calling behavior. Sixth, the ``repair $>$ original''
effect indicates that \method is also patching baseline failures, useful in
deployment, but a complication for clean attribution to drift recovery
specifically. Seventh, strict string-matching against gold calls undercounts true
recoveries: manual inspection of the $13$ unrecovered cases on the three
non-primary models (Section~\ref{sec:error-analysis}) shows that $11$ are
formatting or paraphrase artifacts, so the reported recovery rates should be
read as lower bounds. Finally, while we now report broader cross-protocol
significance for oracle-vs-non-oracle repaired outcomes on the shared Qwen slices,
we still lack a broader wall-clock latency sweep across benchmarks, providers, and
models.

\section{Conclusion}
We studied \method, a lightweight inference-time validation-and-repair stack for
tool calling under interface drift. On Qwen3.5-9B under a non-oracle candidate-retry protocol,
\method improves DICE-300 from $0.717$ to $0.813$ and BFCL-200 from $0.580$ to
$0.805$, with no observed repair harms on the originally-correct slice in both
benchmarks. Three findings qualify the picture. A naive-retry baseline does not
explain the gains. Description and schema drift cause only small degradations on
our DICE slice, while candidate-set drift is the dominant failure mode. And in the
non-oracle setting, missing tool names are the main repair bottleneck: a
candidate-set retry over the visible tools substantially reduces unresolved cases
and closes most of the gap to oracle-constrained upper bounds. Additional
oracle-constrained sweeps across other models and upper-bound ablations remain
useful context, and a BFCL sanity check on GPT-4o-mini suggests that stronger
closed models suffer less drift damage to begin with. A same-protocol BFCL-200
replication on Qwen3.5-35B-A3B ($0.635 \rightarrow 0.845$) materially strengthens
the main claim, but the cleanest successful non-oracle evidence is still
concentrated in the Qwen family. The paper should therefore be read as a
controlled study of drift sensitivity and inference-time repair rather than as a
complete account of production interface drift.

\paragraph{Broader impact.}
The same repair stack that recovers benign drift could in principle mask adversarial
tool injection, where a malicious distractor tool is added to the candidate set.
Practitioners deploying inference-time repair should pair it with provenance checks
on newly added tools.

\paragraph{GenAI disclosure.}
Generative AI tools (Claude, ChatGPT) were used during manuscript preparation for
drafting assistance and code review. All AI-generated content was reviewed, edited,
and verified by the authors, who take full responsibility for the final text.

\bibliographystyle{splncs04}
\bibliography{references}

\begin{thebibliography}{10}
\providecommand{\url}[1]{\texttt{#1}}
\providecommand{\urlprefix}{URL }
\providecommand{\doi}[1]{https://doi.org/#1}

\bibitem{chattopadhyay2026metatuning}
Chattopadhyay, A., Dandekar, R., Roy, K.: Metatuning: An empirical study of
  judge-guided prompt refinement and its boundary conditions. Neurosymbolic
  Artificial Intelligence pp. 1--15 (2026). \doi{10.1177/29498732261443132}

\bibitem{faghih2025gaming}
Faghih, K., Wang, W., Cheng, Y., Bharti, S., Sriramanan, G., Balasubramanian,
  S., Hosseini, P., Feizi, S.: Gaming tool preferences in agentic llms. arXiv
  preprint arXiv:2505.18135  (2025)

\bibitem{geng2023grammar}
Geng, S., Josifoski, M., Peyrard, M., West, R.: Grammar-constrained decoding
  for structured {NLP} tasks without finetuning. In: Proceedings of the 2023
  Conference on Empirical Methods in Natural Language Processing (2023)

\bibitem{guo2026rewrite}
Guo, Y., et~al.: Learning to rewrite tool descriptions for reliable llm-agent
  tool use. arXiv preprint arXiv:2602.20426  (2026)

\bibitem{jang2025dicebench}
Jang, K., Lee, D., Kim, K., Heo, D., Lee, T., Kim, W., Suh, B.: Dice-bench:
  Evaluating the tool-use capabilities of large language models in multi-round,
  multi-party dialogues. arXiv preprint arXiv:2506.22853  (2025)

\bibitem{lercher2025autoguard}
Lercher, A., Bauer, C., Macho, C., Pinzger, M.: Autoguard: Reporting breaking
  changes of rest apis from java spring boot source code. In: 2025 IEEE 32nd
  International Conference on Software Analysis, Evolution and Reengineering
  (SANER) (2025). \doi{10.1109/SANER64311.2025.00083}

\bibitem{li2023apibank}
Li, M., Zhao, Y., Yu, B., Song, F., Li, H., Yu, H., Li, Z., Huang, F., Li, Y.:
  {API}-bank: A comprehensive benchmark for tool-augmented {LLM}s. In:
  Proceedings of the 2023 Conference on Empirical Methods in Natural Language
  Processing. pp. 3102--3116 (2023). \doi{10.18653/v1/2023.emnlp-main.187}

\bibitem{liu2024apigen}
Liu, Z., Hoang, T., Zhang, J., Zhu, M., Lan, T., Kokane, S., Tan, J., Yao, W.,
  Liu, Z., Feng, Y., Murthy, R., Yang, L., Savarese, S., Niebles, J.C., Wang,
  H., Heinecke, S., Xiong, C.: Apigen: Automated pipeline for generating
  verifiable and diverse function-calling datasets. arXiv preprint
  arXiv:2406.18518  (2024)

\bibitem{lu2024toolsandbox}
Lu, J., Holleis, T., Zhang, Y., Aumayer, B., Nan, F., Bai, F., Ma, S., Ma, S.,
  Li, M., Yin, G., Wang, Z., Pang, R.: Toolsandbox: A stateful, conversational,
  interactive evaluation benchmark for {LLM} tool use capabilities. arXiv
  preprint arXiv:2408.04682  (2024)

\bibitem{notionchangelog2026}
{Notion}: Changelog. \url{https://developers.notion.com/page/changelog} (2026),
  notion developer changelog, accessed 2026-04-14

\bibitem{park2024grammar}
Park, K., Wang, J., Berg-Kirkpatrick, T., Polikarpova, N., D'Antoni, L.:
  Grammar-aligned decoding. In: Advances in Neural Information Processing
  Systems (2024)

\bibitem{patil2023gorilla}
Patil, S.G., Zhang, T., Wang, X., Gonzalez, J.E.: Gorilla: Large language model
  connected with massive apis. arXiv preprint arXiv:2305.15334  (2023)

\bibitem{qin2023toolllm}
Qin, Y., Liang, S., Ye, Y., Zhu, K., Yan, L., Lu, Y., Lin, Y., Cong, X., Tang,
  X., Qian, B., Zhao, S., Hong, L., Tian, R., Xie, R., Zhou, J., Gerstein, M.,
  Li, D., Liu, Z., Sun, M.: Toolllm: Facilitating large language models to
  master 16000+ real-world apis. arXiv preprint arXiv:2307.16789  (2023)

\bibitem{rabinovich2025robustness}
Rabinovich, E., Anaby-Tavor, A.: On the robustness of agentic function calling.
  In: Proceedings of the 5th Workshop on Trustworthy NLP (2025)

\bibitem{schick2023toolformer}
Schick, T., Dwivedi-Yu, J., Dessi, R., Raileanu, R., Lomeli, M., Zettlemoyer,
  L., Cancedda, N., Scialom, T.: Toolformer: Language models can teach
  themselves to use tools. arXiv preprint arXiv:2302.04761  (2023)

\bibitem{serbout2023versioning}
Serbout, S., Pautasso, C.: An empirical study of web api versioning practices.
  In: Web Engineering. Lecture Notes in Computer Science (2023).
  \doi{10.1007/978-3-031-34444-2_22}

\bibitem{stripecheckout2022}
{Stripe}: Replaces line item parameters from the create checkout session
  endpoint.
  \url{https://docs.stripe.com/changelog/2022-08-01/replaces-line-item-create-checkout-session}
  (2022), stripe changelog entry, accessed 2026-04-13

\bibitem{stripedefaultvalue2025}
{Stripe}: Adds default value to custom fields on payment links.
  \url{https://docs.stripe.com/changelog/basil/2025-03-31/default-value-payment-links-custom-fields}
  (2025), stripe changelog entry, accessed 2026-04-13

\bibitem{stripedelaydays2025}
{Stripe}: Renames the parameter for custom settlement timing on the balance
  settings api.
  \url{https://docs.stripe.com/changelog/basil/2025-07-30/balance-settings-delay-days-rename}
  (2025), stripe changelog entry, accessed 2026-04-14

\bibitem{stripeinvoicepreview2025}
{Stripe}: Replaces upcoming invoice api methods with the create preview invoice
  api.
  \url{https://docs.stripe.com/changelog/basil/2025-03-31/invoice-preview-api-deprecations}
  (2025), stripe changelog entry, accessed 2026-04-14

\bibitem{bfcl2024}
Yan, F., Mao, H., Ji, C.C.J., Zhang, T., Patil, S.G., Stoica, I., Gonzalez,
  J.E.: Berkeley function calling leaderboard.
  \url{https://gorilla.cs.berkeley.edu/leaderboard.html} (2024)

\bibitem{yao2024taubench}
Yao, S., Shinn, N., Razavi, P., Narasimhan, K.: $\tau$-bench: A benchmark for
  tool-agent-user interaction in real-world domains. arXiv preprint
  arXiv:2406.12045  (2024)

\bibitem{yao2022react}
Yao, S., Zhao, J., Yu, D., Du, N., Shafran, I., Narasimhan, K., Cao, Y.: React:
  Synergizing reasoning and acting in language models. arXiv preprint
  arXiv:2210.03629  (2022)

\bibitem{zhang2026fission}
Zhang, Y., et~al.: Robust tool use via fission-grpo. arXiv preprint
  arXiv:2601.15625  (2026)

\bibitem{zhong2025complexfuncbench}
Zhong, Y., et~al.: Complexfuncbench: Exploring multi-function calling in large
  language models. arXiv preprint arXiv:2501.10132  (2025)

\end{thebibliography}

\end{document}